\documentclass{article}

\usepackage[nonatbib,preprint]{neurips_2019}
\usepackage[utf8]{inputenc} 
\usepackage[T1]{fontenc}    
\usepackage{hyperref}       
\usepackage{url}            
\usepackage{booktabs}       
\usepackage{amsfonts}       
\usepackage{nicefrac}       
\usepackage{microtype}      
\usepackage{mathtools}
\usepackage{amsmath}
\usepackage{multirow}
\usepackage{wrapfig}
\usepackage{pifont}
\usepackage{diagbox}
\usepackage{xcolor}
\usepackage{bbm}

\newcommand{\eg}{\textit{e}.\textit{g}.}
\newcommand{\ie}{\textit{i}.\textit{e}.}

\title{Mobile Video Action Recognition}

\author{
  Yuqi Huo, Xiaoli Xu, Yao Lu, Yulei Niu, Zhiwu Lu, and Ji-Rong Wen\\
  Beijing Key Laboratory of Big Data Management and Analysis Methods\\
  School of Information, Renmin University of China, Beijing 100872, China\\
}

\begin{document}

\maketitle

\begin{abstract}
Video action recognition, which is topical in computer vision and video analysis, aims to allocate a short video clip to a pre-defined category such as brushing hair or climbing stairs. Recent works focus on action recognition with deep neural networks that achieve state-of-the-art results in need of high-performance platforms. Despite the fast development of mobile computing, video action recognition on mobile devices has not been fully discussed. In this paper, we focus on the novel mobile video action recognition task, where only the computational capabilities of mobile devices are accessible. Instead of raw videos with huge storage, we choose to extract multiple modalities (including I-frames, motion vectors, and residuals) directly from compressed videos. By employing MobileNetV2 as backbone, we propose a novel Temporal Trilinear Pooling (TTP) module to fuse the multiple modalities for mobile video action recognition. In addition to motion vectors, we also provide a temporal fusion method to explicitly induce the temporal context. The efficiency test on a mobile device indicates that our model can perform mobile video action recognition at about 40FPS. The comparative results on two benchmarks show that our model outperforms existing action recognition methods in model size and time consuming, but with competitive accuracy.
\end{abstract}

\vspace{-0.1cm}
\section{Introduction}
\label{sect:intro}
\vspace{-0.1cm}

Video analysis has drawn increasing attention from the computer vision community, for that videos occupy more than 75\% of the global IP traffic~\cite{index2017forecast}. With the development of deep learning methods, recent works achieve promising performance on video analysis tasks, including action recognition~\cite{feichtenhofer2016convolutional, peng2016multi, simonyan2014two, wang2016temporal, NIPS2018_7489}, emotion recognition~\cite{xu2018heterogeneous}, human detection~\cite{NIPS2018_7988}, and deception detection~\cite{perez2015verbal}. Taking video action recognition as example, most works focused on raw video analysis using deep learning models and optical flow~\cite{simonyan2014two, wang2016temporal} without limit of storage and computation. Furthermore, compressed video action recognition~\cite{wu2018compressed} is proposed to replace raw videos (\ie, RGB frames) with compressed representations (\eg, MPEG-4), which retain only a few key frames and their offsets (\ie, motion vectors and residual errors) for storage reduction. However, none of these works can be directly applied on mobile devices, which have limited storage for both video data and analysis model. Therefore, the next-generation video analysis technique on mobile devices is expected to satisfy that 1) the framework is lightweight; 2) the model is able to deal with compressed videos.

In this paper, we present a novel video analysis task, called \textit{mobile video action recognition}, to fill in the above gap in video analysis (also see Table \ref{tab:overview}). Specifically, mobile video action recognition aims to perform efficient action recognition with compressed videos on mobile devices, considering the limit of storage and computation. This novel task meets the tendency of video analysis on mobile phones. On one hand, action recognition can help to automatically tag users' videos on social networking apps (\eg, Facebook and Instagram) before uploading from mobile phones. On the other hand, conducting video analysis on mobile devices can reduce the overload and computation on the cloud server. The key challenges of mobile video action recognition are: 1) how to design a lightweight and high-performance framework applicable on mobile devices; 2) how to extract meaningful representations from compressed videos. Thanks to the rapid development of neural network compression~\cite{liu2018mobile, Sandler_2018_CVPR, Wu_2016_CVPR, Yang_2018_ECCV}, the compressed small networks can be used as the backbone for overcoming the first challenge, which balance the effectiveness and efficiency. In addition, compressed videos (\eg, MPEG-4) have been exploited~\cite{zhang2018real} to replace raw videos for saving storage: multiple modalities (including RGB I-frame (\textbf{I}), low-resolution motion vector (\textbf{MV}) and residual (\textbf{R})~\cite{wu2018compressed}) are extracted to avoid bringing in extra optical flow data. However,~\cite{wu2018compressed} used three deep models to process I-frames, motion vectors and residuals, and simply summed prediction scores from all these modalities in an ensemble manner. We argue that the ensemble strategy cannot fully capture the inner interactions between different modalities from compressed videos.

\begin{table}[t]
  \centering
  \caption{Overview of state-of-the-art video analysis models. ``Compressed'' denotes the usage of compressed videos. ``Mobile'' denotes whether a model is applicable on mobile devices.}
  \label{tab:overview}
  \vspace{-0.05in}
  \tabcolsep7pt
    \begin{tabular}{|c|c|c|}
    \hline
    \diagbox{Compressed}{Mobile} & No & Yes \\
    \hline
    \multirow{2}{*}{No} &  C3D~\cite{tran2015learning}, Two-Stream~\cite{simonyan2014two} & \multirow{2}{*}{Bottleneck-LSTM~\cite{liu2018mobile}} \\
     & TSN~\cite{wang2016temporal}, TS R-CNN~\cite{peng2016multi} & \\
    \hline
    \multirow{2}{*}{Yes} & \multirow{2}{*}{ CoViAR~\cite{wu2018compressed}, DTMV-CNN~\cite{zhang2018real}} & \multirow{2}{*}{\textbf{TTP (Ours)} } \\
    & & \\
    \hline
    \end{tabular}
    \vspace{-0.1in}
\end{table}
\begin{wrapfigure}{r}{0cm}
  \centering
  \vspace{-0.1in}
  \includegraphics[width=0.5\columnwidth]{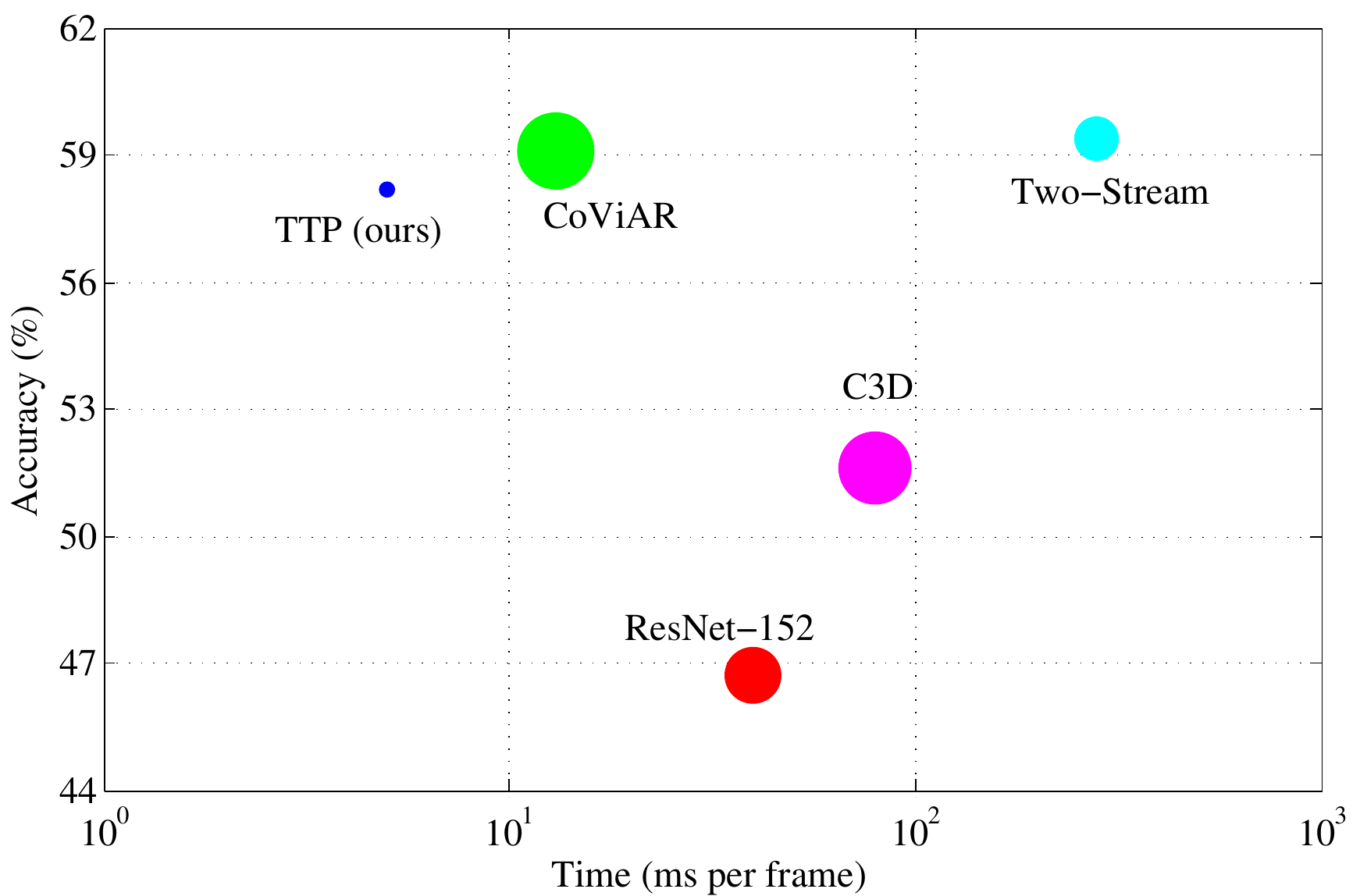}
  \vspace{-0.1in}
  \caption{Efficiency and accuracy comparison on benchmark HMDB-51~\cite{kuehne2011hmdb} with the same platform. Representative baselines include CoViAR~\cite{wu2018compressed}, two-stream network~\cite{simonyan2014two}, C3D~\cite{feichtenhofer2016convolutional}, and ResNet-152~\cite{he2016deep}. The node size denotes the scale (\ie, parameters) of the corresponding model.}
  \label{fig:exp}
\end{wrapfigure}

According to the above considerations, we propose a lightweight framework to solve the mobile video action recognition task. Specifically, we employ MobileNetV2~\cite{Sandler_2018_CVPR} as the backbone network to process the multiple modalities (including RGB I-frame \textbf{I}, motion vectors \textbf{MV}, and residual \textbf{R}) extracted from compressed videos. Instead of the score ensemble strategy used in~\cite{wu2018compressed}, we further propose a novel Temporal Trilinear Pooling (TTP) for fusing the extracted multiple modalities. In addition to motion vectors, we also put forward a temporal fusion method by combining successive key frames to achieve better temporal representation. As shown in Table \ref{tab:overview}, the proposed TTP does fill in the gap in video analysis. Particularly, the efficiency test on a mobile device indicates that our TTP can perform mobile video action recognition at about 40FPS (see Table~\ref{tab:speed}). Moreover, as shown in Figure~\ref{fig:exp}, our TTP achieves competitive performance but with extremely fewer parameters, as compared to state-of-the-art methods for video action recognition. This observation is further supported by the extensive results reported on two action recognition benchmarks (see Table~\ref{tab:result}).

Our contributions are: (1) We present a novel mobile video action recognition task, which aims to perform efficient action recognition with compressed videos on mobile devices. This novel task fills in the gap in action recognition and video analysis. (2) We propose a lightweight framework with a trilinear pooling module for fusing the multiple modalities extracted from compressed videos. (3) In addition to the motion vector information used for replacing optical flow, we provide a temporal fusion method to explicitly induce the temporal context into mobile video action recognition.

\vspace{-0.1cm}
\section{Related Work}
\label{sect:related}
\vspace{-0.1cm}

\paragraph{Video Action Recognition}
Before the emergence of deep learning techniques, handcrafted features including 3D-HOG~\cite{klaser2008spatio} and STIP~\cite{laptev2005space} are extracted from videos to solve the action recognition task. They need to be combined with optical flow to represent the temporal structure. The presence of convolutional neural network (CNN)~\cite{krizhevsky2012imagenet} promotes the feature representation significantly, and two-stream network~\cite{simonyan2014two} is proposed to utilize two CNNs to model raw video frames and optical flow separately. Various improved versions such as temporal segment network (TSN)~\cite{wang2016temporal} are designed to capture the long-range temporal structure, but they still need to fuse optical flow. C3D~\cite{tran2015learning} is proposed to model the temporal structure with one stream 3D CNN, where it avoids using optical flow but still costs too much for bringing in 3D convolution operation. Note that videos could not be stored in raw version in mobile devices, and they need to be transformed and stored in compression format such as MPEG-4, H.264, etc. Above approaches all process raw videos (but not compressed ones), and they have too many parameters or large input data size. Therefore, none of them could be implemented on a resource-limited mobile device. ~\cite{zhang2018real} tries to extract motion vectors from compressed videos to simulate optical flow, but the raw RGB frames are still used for action recognition. Recently,~\cite{wu2018compressed} takes only compressed videos as input, but ResNet-152~\cite{he2016deep} is used as the core CNN module, which has too many parameters to be implemented on mobile devices. Moreover,~\cite{wu2018compressed} regards multiple modalities separately and the whole training and testing process is not end-to-end.

\paragraph{Pooling Methods}
Pooling methods are requisite either in two-stream networks~\cite{simonyan2014two, wang2016temporal} or in other feature fusion models.~\cite{wang2016temporal} simply uses average pooling and outperforms others.~\cite{lin2015bilinear} proposes bilinear pooling to model local parts of object: two feature representations are learned separately and then multiplied using the outer product to obtain the holistic representation.~\cite{wang2017spatiotemporal} combines two-stream network with a compact bilinear representation~\cite{gao2016compact}.~\cite{cui2017kernel} defines a general kernel-based pooling framework which captures higher-order interactions of features. However, most existing bilinear pooling models are capable to combine only two features, and none of their variants could cope with more than two features, which is needed in video action recognition.

\paragraph{Lightweight Neural Networks}
Recently, lightweight neural networks including SqeezeNet~\cite{iandola2016squeezenet}, Xception~\cite{chollet2017xception}, ShuffleNet~\cite{zhang2018shufflenet}, ShuffleNetV2~\cite{ma2018shufflenet}, MobileNet~\cite{howard2017mobilenets}, and MobileNetV2~\cite{Sandler_2018_CVPR} have been proposed to run on mobile devices with the parameters and computation reduced significantly. Since we focus on mobile video action recognition, all these lightweight models could be use as backbone.

\section{Problem Definition}

We have discussed the background of video action recognition in Sections~\ref{sect:intro} and \ref{sect:related}. In this section, we first give the details of extracting multiple modalities from compressed videos for action recognition, and then define our mobile video action recognition problem with the extracted modalities.

\subsection{Extracting Multiple Modalities from Compressed Videos}

We follow~\cite{le1991mpeg,wu2018compressed} for the usage of compressed videos in action recognition. Formally, there are several segments in a compressed video. Given one video segment with $n$ frames, we treat all frames as a sequence $F = [f_0, f_1, \dots, f_{n-1}]$, where each frame $f_i \in \mathbb{R}^{h \times w \times 3}$ is a RGB image with 3 channels, and $h$ and $w$ represent the height and width respectively. Since video compression leverages the spatial continuity of adjacent frames, only some key frames instead of all frames need to be stored. Against storing all frames ($[f_0, f_1, \dots, f_{n-1}]$) in uncompressed videos, compressed videos only store $f_0$ as the intra-coded frame in this segment, which is called I-frame (\textbf{I}). The rest frames in this segment are treated as P-frames, where the differences between each P-frame $f_i$ and the I-frame $f_0$ are extracted and stored as motion vector (\textbf{MV}) and residual information (\textbf{R}). Specifically, the motion information is described as ``motion vector'' and stored in P-frames as $\textbf{MV}\!=\![\textrm{MV}_1, \textrm{MV}_2, \dots, \textrm{MV}_{n-1}]$ where $\textrm{MV}_i\!\in\!\mathbb{R}^{h \times w \times 2}$. Note that motion vectors have only 2 channels because the motion information is described in height and width channels but not color channels. For each position $(mv_{ix}, mv_{iy})$ in $\textrm{MV}_i$, $mv_{ix}$ and $mv_{iy}$ depict the horizontal and vertical movements from frame $f_0$ to frame $f_i$, where $x$ and $y$ are the coordinates of the grid. However, the motion vector itself is not enough to store all the information in $f_i$, and there still exists some small differentiation. The differentiation between I-frame and the I-frame plus motion vector is stored as residual information $\textbf{R}\!=\![\textrm{R}_1, \textrm{R}_2, \dots, \textrm{R}_{n-1}]$, where $\textrm{R}_i \!\in\!\mathbb{R}^{h \times w \times 3}$ is an RGB-like image. Therefore, $F$ is transformed to $F^{'}\!=\![f^{'}_0, f^{'}_1, \dots, f^{'}_{n-1}]$, where each frame $f^{'}_i$ is formalized as
\begin{equation}
f^{'}_i = f^{'}_i(x, y, z) =
\begin{cases}
f_0(x, y, z) & {i = 0}\\
f_0(x + mv_{ix}, y + mv_{iy}, z) + r_i(x, y, z) & {i \ne 0}
\end{cases}.
\end{equation}
where $x$, $y$, $z$ denote the coordinates in the three dimensions. Note that each video segment generates one I-frame and several P-frames. In MPEG-4 video, the segment length $n$ equals 12 in average, so that each I-frame has their individual 11 P-frames. Since the compressed video is the universal format stored in mobile devices, we can extract motion vectors directly from compression videos to simulate the optical flow, which could assist to construct the temporal structure for action recognition.

\subsection{Mobile Video Action Recognition with Multiple Modalities }

We formally define our mobile video action recognition problem as follows. Given the three modalities (i.e. \textbf{I}, \textbf{MV}, and \textbf{R}) extracted from a compressed video, the goal of mobile video action recognition is to perform efficient action recognition (\ie, aligning the unique action category with the video) on mobile devices. Recent works utilize different CNNs to process the modalities independently. In~\cite{wu2018compressed}, Resnet-152~\cite{he2016deep} is used to process $\textbf{I}$, which is effective but consumes too much time and storage (see Table~\ref{fig:exp}). In this work, we uniformly adopt MobileNetV2~\cite{Sandler_2018_CVPR} as backbone. Specifically, $\textrm{MoblieNet}_I$ takes $\textbf{I} \in \mathbb{R}^{h \times w \times 3}$ as input, while $\textrm{MoblieNet}_{MV}$ takes $\textbf{MV}$ and $\textrm{MoblieNet}_R$ takes $\textbf{R}$ as input separately. In~\cite{wu2018compressed}, the three modalities are only late fused at the test phase, and each modality is processed independently during the training stage. Therefore, the interactions between different modalities are not fully explored, and the multi-modal fusion is not consistent between the training and test stages, which limits the performance of action recognition.

\begin{figure}[t]
  \centering
  \includegraphics[width=0.92\columnwidth]{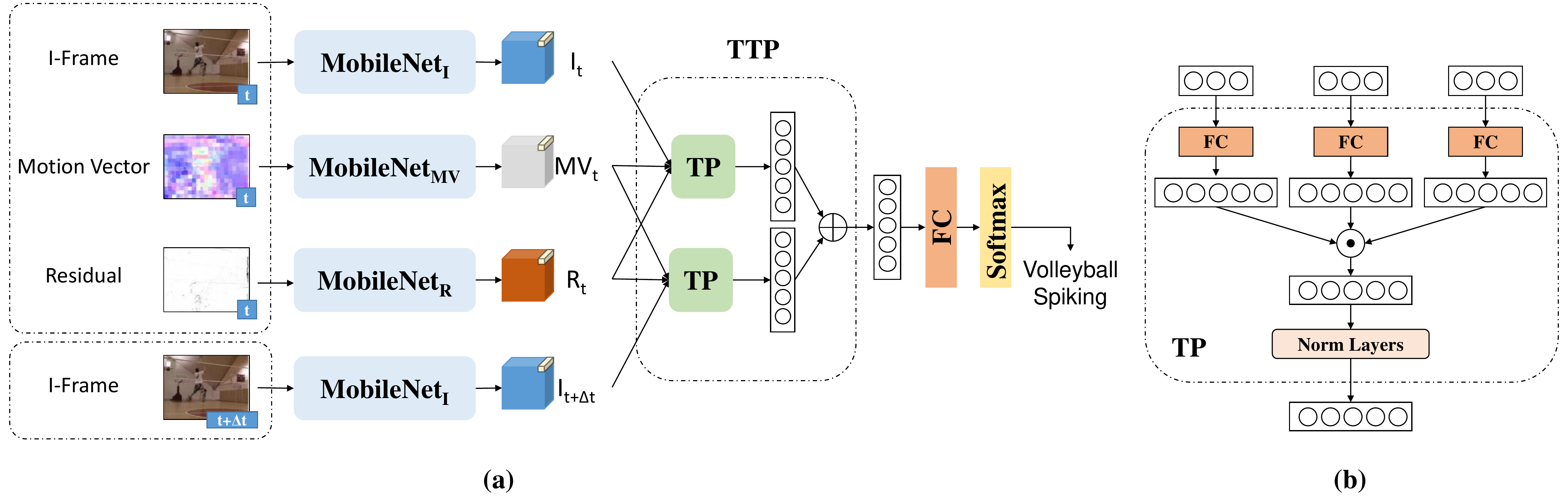}
  \caption{(a) Overview of network architecture of Temporal Trilinear Pooling (TTP). (b) Complete design of the Trilinear Pooling (TP) module. }
  \label{fig:framework}
  \vspace{-0.1in}
\end{figure}

\section{Methodology}

In this section, we first introduce the trilinear pooling method for multi-modal fusion in mobile video action recognition. We then refine the multi-modal representation by inducing the temporal context information. The overall architecture is illustrated in Figure~\ref{fig:framework}(a).

\subsection{Trilinear Pooling}

In this work, we propose a novel trilinear pooling module to model three factor variations together. This is motivated by bilinear models~\cite{lin2015bilinear,yu2017multi}, which are initially proposed to model two-factor variations, such as ``style'' and ``content''. For mobile video action recognition, we generalize the bilinear pooling method to fuse the three modalities extracted from compressed videos.

Specifically, a feature vector (usually deep one) is denoted as $x \in \mathbb{R}^c$, where $c$ is the dimensions of the feature $x$. The bilinear combination of two feature vectors with the same dimension $x \in \mathbb{R}^{c}$ and $y \in \mathbb{R}^{c}$ is defined as $xy^\mathrm{T} \in \mathbb{R}^{c \times c}$~\cite{lin2015bilinear}. In general, given two representation matrices $X= [x_1;x_2;\cdots;x_K] \in \mathbb{R}^{c \times K}$ and $Y= [y_1;y_2;\cdots;y_{K}] \in \mathbb{R}^{c \times K}$ for two frames, a pooling layer follows the bilinear combination of these two matrices:
\begin{equation}
f_{\textrm{BP}}(X,Y) = \frac{1}{K}\sum_{i=1}^{K}x_iy^\mathrm{T}_i = \frac{1}{K}XY^\mathrm{T}.\label{eq:gram}
\end{equation}

It can be clearly seen that the above bilinear pooling allows the outputs ($X$ and $Y$) of the feature extractor to be conditioned on each other by considering all their pairwise interactions in form of a quadratic kernel expansion. However, this results in very high-dimensional features with a large number of parameters involved. To address this problem, Multi-modal Factorized Bilinear (MFB)~\cite{yu2017multi} introduces an efficient attention mechanism into the original bilinear pooling based on the Hadamard product. The full MFB model is defined as follows:
\begin{equation}
f_{\textrm{MFB}}(x,y)_i = x^\mathrm{T}F_iy\label{eq:mfb},
\end{equation}
where $F= [F_1;F_2; \cdots;F_D] \in \mathbb{R}^{c\times c \times D}$, $D$ is the number of projection matrices, and each $x^\mathrm{T}F_{i}y$ contributes to one dimension of the resultant feature. Obviously, the MFB model reduces the output dimension from $c \times c$ to $D$, which causes much less computational cost. According to the matrix factorization technique~\cite{rendle2010factorization}, each $F_i$ in the $F$ can be factorized into two low-rank matrices:
\begin{equation}
f_{\textrm{MFB}}(x,y)_i = x^\mathrm{T}F_iy = x^\mathrm{T}U_iV^\mathrm{T}_iy = \sum_{j=1}^{d}x^\mathrm{T}u_{ij}v^\mathrm{T}_{ij}y = \mathbbm{1}^\mathrm{T}(U^\mathrm{T}_ix \odot V^\mathrm{T}_iy)
\label{eq:mfb_final},
\end{equation}
where $U_i \in \mathbb{R}^{c \times d}$ and $V_i \in \mathbb{R}^{c \times d}$ are projection matrices, $\odot$ is the Hadamard product, $\mathbbm{1} \in \mathbb{R}^{d}$ is an all-one vector, and $d$ denotes the dimension of these factorized matrices. Therefore, we only need to learn $U = [U_1;U_2;\cdots;U_D] \in \mathbb{R}^{c \times d \times D}$ and $V = [V_1;V_2;\cdots;V_D] \in \mathbb{R}^{c\times d \times D}$.

Inspired by Eq.(\ref{eq:mfb_final}), we propose a novel trilinear pooling method, which aims to fuse three feature vectors ($x$, $y$ and $z$). Unlike bilinear pooling that can combine only two feature vectors, our Trilinear Pooling method fuse $x$, $y$ and $z$ using the Hadamard product:
\begin{equation}
f_{\textrm{TP}}(x,y,z) = \mathbbm{1}^\mathrm{T}(U^\mathrm{T}x \odot V^\mathrm{T}y \odot W^\mathrm{T}z),\label{eq:TLP}
\end{equation}
where $W$ is also a projection matrix $W = [W_1;W_2;\cdots;W_D] \in \mathbb{R}^{c\times d \times D}$, and $f_{\textrm{TP}}$ denotes the output of trilinear pooling. Note that our proposed trilinear pooling degrades to MFB if all elements in $W$ and $z$ are fixed as 1. When the inputs are generalized to feature maps (\ie, $X=[x_i], Y=[y_i], Z=[z_i] \in \mathbb{R}^{c \times K}$), every position of these feature maps makes up one group of inputs, and the outputs of them are summed element-wised (sum pooling) as:
\begin{equation}
f_{\textrm{TP}}(X,Y,Z) = \sum_{i=1}^{K}f_{\textrm{TP}}(x_i,y_i,z_i).\label{eq:TLP_final}
\end{equation}
In this paper, we utilize the Trilinear Pooling model to obtain the multi-modal representation of $t$-th segment by fusing I-frame $\textrm{I}_t$, motion vector $\textrm{MV}_t$ and residual $\textrm{R}_t$ (see Figure~\ref{fig:framework}(b)):
\begin{equation}
f_{\textrm{TP}}(\textrm{I}_t, \textrm{MV}_t, \textrm{R}_t) =
\sum_{i=1}^{K}f_{\textrm{TP}}(\textrm{I}_{t,i},\textrm{MV}_{t,i},\textrm{R}_{t,i}),\label{eq:TP}
\end{equation}
where $\textrm{I}_t$, $\textrm{MV}_t$ and $\textrm{R}_t$ are the output feature maps from penultimate layer of $\textrm{MobileNet}_{I}$, $\textrm{MobileNet}_{MV}$ and $\textrm{MobileNet}_{R}$, respectively. For each segment $t$, the only I-frame is selected as $\textrm{I}_t$, while one $\textrm{MV}_t$ and one $\textrm{R}_t$ are randomly selected. As in~\cite{kim2016hadamard}, the trilinear vector is then processed with a signed square root step ($f\leftarrow \textrm{sign}(f)\sqrt{|f|}$), followed by $l_2$ normalization ($f\leftarrow f/||f||$).

\subsection{Temporal Trilinear Pooling}

Motion vector is initially introduced to represent the temporal structure as optical flow does. However, compared to the high-resolution optical flow, motion vector is so coarse that it only describes the movement of blocks, and thus all values within the same macro-block surrounded by $(mv_{ix}, mv_{iy})$ are identical. Although we have proposed trilinear pooling to address this drawback, the temporal information needs to be explicitly explored. Since residuals are also computed as the difference between the their I-frame, they are strongly correlated with motion vectors. Therefore, we treat the motion vectors and the residuals integrally. Note that the fusion of $\textrm{I}_t$, $\textrm{MV}_t$ and $\textrm{R}_t$ within only one segment is not enough to capture the temporal information. We further choose to include the adjacent segment's information. Specifically, in addition to calculating $f_\textrm{TP}(\textrm{I}_t, \textrm{MV}_t, \textrm{R}_t)$ by trilinear pooling, we also combine $\textrm{MV}_t$ and $\textrm{R}_t$ with $\textrm{I}_{t + \Delta t}$ (\ie~the I-frame in the adjacent segment). The output of temporal trilinear pooling is defined as (see Figure~\ref{fig:framework}(a)):
\begin{equation}
f_\textrm{TTP}(t) = f_\textrm{TP}(t,0) + f_\textrm{TP}(t,\Delta t)
\end{equation}
where $f_\textrm{TP}(t, \Delta t)$ denotes $f_{\textrm{TP}}(\textrm{I}_{t+\Delta t}, \textrm{MV}_t, \textrm{R}_t)$ for simplicity. In this paper, we sample the offset ${\Delta t}$ from $\{-1,1\}$ during the training stage. During the test stage, $\Delta t$ is fixed as $1$ for the first frame and $-1$ for other frames. This temporal fusion method solves the temporal representation drawback without introducing extra parameters, which is efficient and can be implemented on mobile devices. The TTP representation is further put into a fully connected layer to calculate the classification scores $\textbf{s}(t)=P^Tf_{\textrm{TTP}}(t)$, where $P\in\mathbb{R}^{D\times C}$ is learnable parameters and $C$ is the number of categories.

Since mobile video action recognition can be regarded as a multi-class classification problem, we utilize the standard cross-entropy loss for model training:
\begin{equation}
\mathcal{L}_{TTP}(t) = -\log\textrm{softmax}(\textbf{s}_{gt}(t))
\end{equation}
where $\textbf{s}_{gt}(t)$ is the predicted score for $t$-th segment with respect to its ground-truth class label.

\section{Experiments}

\subsection{Datasets}

In this paper, we report results on two widely used benchmark dataset, including HMDB-51~\cite{kuehne2011hmdb} and UCF-101~\cite{soomro2012ucf101}. HMDB-51 contains 6,766 videos from 51 action categories, while UCF-101 contains 13,320 videos from 101 action categories. Both datasets have 3 officially given training/test splits. In HMDB-51, each training/test split contains 3,570 training clips and 1,530 testing clips. In UCF-101, each training/test split contains different number of clips, approximately 9,600 clips in the training split and 3,700 clips in the test split. Since each video in these two datasets is a short clip belonging to a single category, we employ top-1 accuracy on video-level class prediction as the evaluation metric.

\subsection{Implementation Details}

As in~\cite{wang2016temporal, wu2018compressed}, we resize frames in all videos to $340\times256$. In order to implement our model on mobile devices, we simply choose three MobileNetV2~\cite{Sandler_2018_CVPR} pretrained on ImageNet as the core CNN module to extract the representations of I-frames, motion vectors and residuals. All the parameters of the projection layers are randomly initialized. The raw videos are encoded in MPEG-4 format.

In the training phase, in addition to picking 3 segments randomly in each video clip, we also pick three adjacent I-frames for each selected segment for the proposed temporal fusion method. We apply color jittering, horizontally flipping and random cropping to $224\times 224$ for data augmentation, as in previous works. There are two stages in training phase. Firstly, following~\cite{wu2018compressed}, we fine-tune the three CNN modules by using the I-frames, motion vectors and residuals data independently. Secondly, we jointly fine-tune the CNN modules and the TTP module. The two training stages are done with the Adam optimizer: the learning rate is set as 0.01 in the first stage and 0.005 in the second stage, multiplied by 0.1 on plateau. According to~\cite{yu2017multi}, the dimension of projection layers is empirically set to 8,192. In the test phase, we also follow~\cite{wu2018compressed} by randomly sampling 25 segments and their adjacent I-frames for each video. We apply horizontal flips and 5 random crops for data augmentation.

All experiments are implemented with PyTorch. In limited-resource simulation for mobile video action recognition, we select the Nvidia Jetson TX2 platform with 8G memory, 32G storage and a Nvidia Pascal GPU. Moreover, in sufficient-resource simulation for conventional video action recognition, we select the Dell R730 platform with Nvidia Titan Xp GPU.

\subsection{Efficiency Test}

We firstly demonstrate the per-frame running time and FPS of our model in both limited-resource and sufficient-resource environments. We also make comparison to the state-of-the-art CoViAR~\cite{wu2018compressed}, since it also exploits compressed videos and does not use the optical flow information. However, CoViAR utilizes ResNet-152 for I-frames and two ResNet-18 for motion vectors and residual independently, which occupy too much space and are slow to run.

\begin{table}[t]
  \centering
  \caption{Comparison of per-frame inference efficiency. Following CoViAR~\cite{wu2018compressed}, we forward multiple CNNs concurrently. CoViAR cannot run on Jetson TX2 due to out of memory (OOM). }
  \label{tab:speed}
  \vspace{-0.05in}
  \tabcolsep4.5pt
  \begin{tabular}{|c|c|cc|cc|}
  \hline
  \multicolumn{2}{|c|}{} & Ours (Jetson) & CoViAR (Jetson) &  Ours (R730) & CoViAR (R730) \\
  \hline
  \multirow{2}{*}{Time(ms)} & Preprocess & \bf 12.2 & OOM  & \bf 0.6 & 7.8 \\
  & CNN & \bf 24.6 & OOM & \bf 4.3 & 5.1 \\
  \hline
  \multirow{2}{*}{FPS} & Preprocess & \bf 82.1 & OOM  & \bf 1587.3 & 127.6 \\
  & CNN & \bf 40.7 & OOM  & \bf 233.6 & 194.9 \\
  \hline
  \end{tabular}
\end{table}

We compare the efficiency of the two models (\ie~CoViAR and our model) under exactly the same setting. On the Dell R730 platform, the preprocessing phase (including loading networks) is mainly run on two Intel Xeon Silver 4110 CPUs, and the CNN forwarding phase (including extracting motion vectors and residuals) is mainly run on one TITAN Xp GPU. As shown in Table \ref{tab:speed}, our TTP model runs faster than CoViAR in both pre-processing and CNN phases on the Dell R730 platform. The reason is that CoViAR has three large networks with lots of parameters, resulting in huge cost on loading networks. On the contrary, the CNN running time of our method does not overwhelm mainly because the trilinear pooling in our framework costs extra computational time.
Moreover, for efficiency test on the mobile device, the experiments are conducted on the Nvidia Jetson TX2 platform. The preprocessing phase is run on Dual-core Denver 2 64-bit CPU and the CNN forward phase is run on the GPU. The results demonstrate that CoViAR is too storage-consuming to be directly implemented on this device, while our TTP framework fits well in the embedded environment and runs very fast. Note that our model outperforms CoViAR in model size and time consuming, but with competitive accuracy. More supporting results can be found in Table~\ref{tab:result}.

\subsection{Ablation Study}

\begin{table}[t]
  \centering
  \caption{Ablative results (\%) for our TTP model on the two benchmarks (each has three splits).}
  \label{tab:ablation}
  \vspace{-0.05in}
  \tabcolsep10pt
  \begin{tabular}{|c|c|ccccccc|}
  \hline
   Dataset & Splits & I & MV & R & I+MV+R & BP & TP & TTP \\
  \hline
  & Split 1 & 47.7 & 42.8 & 47.7 & 58.5  & 58.6 & 59.3 & \bf 59.7 \\
  & Split 2 & 43.9 & 40.2 & 44.3 & 53.3  & 53.9 & 56.3 & \bf 57.0 \\
  HMDB & Split 3 & 43.9 & 41.8 & 47.8 & 55.1 & 56.3 & 56.6 & \bf 57.9 \\
  \cline{2-9}
  & Avg. & 45.2 & 41.6 & 46.6 & 55.6 & 56.3 & 57.4 & \bf 58.2 \\
  \hline
  & Split 1 & 79.8 & 69.5 & 80.4 & 85.8 & 85.5 & 86.9 & \bf 87.0 \\
  & Split 2 & 79.2 & 69.9 & 79.5 & 85.3 & 86.1 & 86.6 & \bf 87.0 \\
  UCF & Split 3 & 79.6 & 72.5 & 81.3 & 86.4 & 86.5 & 87.6 & \bf 87.7 \\
  \cline{2-9}
  & Avg. & 79.5 & 70.6 & 80.4 & 85.8 & 86.0 & 87.0 & \bf 87.2 \\
  \hline
  \end{tabular}
\end{table}

We conduct experiments to show the benefits of using our TTP model compared with single modality and other fusion options applicable on mobile devices. Specifically, we uniformly use three MobileNetV2 networks to process the three components (\textbf{I}, \textbf{MV} and \textbf{R}) extracted from compressed videos, and demonstrate all the ablative results on the two benchmarks by training with different parts of our TTP model. For single-modality based models, ``I'', ``MV'' and ``R'' denote the results obtained by using $\textrm{MobileNet}_I$, $\textrm{MobileNet}_{MV}$ and $\textrm{MobileNet}_R$, respectively. For late-fusion based model, ``I+MV+R'' indicates that the output is fused by simply adding the score of the three CNNs together. We also compare our TTP model with existing bilinear pooling models, which are the degraded forms of our Trilinear Pooling. Since existing bilinear pooling methods cannot be directly adapted to the three modalities, we simply apply pairwise combination over them, and sum the three predicted classification scores together like ``I+MV+R''. Since conventional bilinear pooling~\cite{lin2015bilinear} and factorized bilinear pooling~\cite{kim2016hadamard} have too many parameters to be implemented on mobile devices, we resort to compact bilinear pooling~\cite{gao2016compact} (denoted as ``BP''), which has much fewer parameters. Finally, ``TP'' is obtained by implementing our Trilinear Pooling module, while ``TTP'' denotes our proposed Temporal Trilinear Pooling module.

As shown in Table \ref{tab:ablation}, single-modality based models (\ie, I, MV, or R) could not achieve good performance without using multi-modal information, which indicates that the compressed video needs to be fully explored in order to obtain high accuracy. I and R yield similar result because they both contain the RGB data: the I-frames contain a small number of informative frames, while the residuals contain a large number of less informative frames. Since the motion vectors only contain the motion information, MV could not perform as well as the other two for action recognition.

For multi-modal fusion, all bilinear/trilinear pooling methods outperform I+MV+R, showing the power of pooling methods instead of linearly late fusion. Moreover, our TP method achieves \%1 improvement over BP, which validates the effectiveness of our proposed pooling method. In addition, TTP outperforms TP, demonstrating the importance of the temporal information.

\subsection{Comparative Results}

We make comprehensive comparison between our TTP method and other state-of-the-art action recognition methods. To this end, we compute the efficiency (\ie, parameters and GFLOPS) and top-1 accuracy as the evaluation metrics for action recognition. In our experiments, all compared methods can be divided into two groups: 1) \textbf{Raw-Video Based Methods}: LRCN~\cite{donahue2015long} and Composite LSTM~\cite{srivastava2015unsupervised} utilize RNNs to process the optical flow information, while Two-stream~\cite{simonyan2014two} adopts two-way CNNs to process the optical flow information. C3D~\cite{tran2015learning}, TSN (RGB-only)~\cite{wang2016temporal} and ResNet~\cite{he2016deep} employ large CNN models over RGB frames without using other information. 2) \textbf{Compressed-Video Based Methods}: DTMV-CNN~\cite{zhang2018real} integrates both compressed videos and raw ones into two-stream network, while CoViAR~\cite{wu2018compressed} is the most closely related model (w.r.t. our TTP) that replaces raw videos by compressed ones, without using any optical flow for action recognition.

The comparative results are shown in Table~\ref{tab:result}. We have the following observations: (1) Our TTP model is the most efficient for video action recognition. Specifically, it contains only 17.5 $\times10^6$ parameters and has only average 1.4 GFLOPs over all frames. (2) Our TTP model outperforms most of state-of-the-art methods according to both efficiency and accuracy. (3) Although our TTP model achieves slightly lower accuracies than Two-stream, CoViAR and DTMV-CNN, it leads to significant efficiency improvements over these three methods. (4) As compared to CoViAR, our TTP model saves nearly 80\% of the storage. Note that ``I+MV+R'' in Table~\ref{tab:ablation} is essentially CoViAR by using MobileNetV2 as the core CNN module. Under such fair comparison setting, our TTP model consistently yields accuracy improvements over CoViAR on most of the dataset splits. (5) Compressed-video based models generally competitive accuracies compared to raw-video based models, showing the high feasibility of exploring compressed video in action recognition.

\begin{table}[t]
  \centering
  \caption{Comparative results on the two benchmark datasets for both raw-video based methods and compressed-video based ones. ``CV.'' denotes the usage of compressed videos for action recognition. ``OF.'' denotes the usage of optical flow. ${\ddagger}$ indicates that the model only uses RGB frames. }
  \label{tab:result}
  \vspace{-0.05in}
  \tabcolsep8pt
  \begin{tabular}{|l|cc|cc|cc|}
  \hline
     & \multicolumn{2}{c|}{Setting} &  \multicolumn{2}{c|}{Efficiency}  &  \multicolumn{2}{c|}{Accuracy} \\
  \hline
  Model   & CV. & OF. & Param. ($\times10^6$) & GFLOPs & HMDB  & UCF \\
  \hline
  LRCN~\cite{donahue2015long} & N & Y & 114.8 & 15.5 & -- & 82.7 \\
  Composite LSTM~\cite{srivastava2015unsupervised} & N & Y & -- & -- & 44.0 & 84.3 \\
  Two-Stream~\cite{simonyan2014two} & N & Y & 46.6 & 3.3 & \bf59.4 & 88.0 \\
  C3D~\cite{tran2015learning} & N & N & 78.4 & 38.5 & 51.6 & 82.3 \\
  TSN$^{\ddagger}$~\cite{wang2016temporal} & N & N & 31.2 & 3.8 & -- & 85.7 \\
  ResNet-50~\cite{he2016deep} & N & N & 25.6 & 3.8 & 46.7 & 82.3 \\
  ResNet-152~\cite{he2016deep} & N & N & 60.2 & 11.3 & 48.9 & 83.4 \\
  \hline
  DTMV-CNN~\cite{zhang2018real} & Both & N & 46.6 & 1.9 & 55.3 & 87.5 \\
  CoViAR~\cite{wu2018compressed} & Y & N & 83.6 & 14.9 & 59.1 & \bf90.4 \\
  \hline
  TTP (ours)& Y & N & \bf17.5 & \bf1.4 & 58.2 & 87.2 \\
  \hline
  \end{tabular}
\end{table}

\section{Conclusion}

Video action recognition is a hot topic on computer vision. However, in the mobile computing age, there lacks the exploration of this task on mobile devices. In this paper, we thus focus on a novel mobile video action recognition task, where only the computational capabilities of mobile devices are accessible. Instead of raw videos with huge storage, we choose to extract multiple modalities (including I-frames, motion vectors, and residuals) directly from compressed videos. By employing MobileNetV2 as the backbone network, we propose a novel Temporal Trilinear Pooling (TTP) module to fuse the multiple modalities for mobile video action recognition. In addition to motion vectors, we also provide a temporal fusion method by combining successive key frames to achieve better temporal representation. The efficiency test on a mobile device indicates that our TTP model can perform mobile video action recognition at about 40FPS. The comparative results on two benchmarks demonstrate that our TTP model outperforms existing action recognition methods in model size and time consuming, but with competitive accuracy. In our ongoing research, we will introduce the attention strategy into mobile video action recognition to obtain better results.


\end{document}